\documentclass[a4paper,conference]{IEEEtran}
\usepackage{amsmath}
\usepackage{amssymb}
\usepackage{color}
\usepackage{graphicx}
\usepackage{subfig}
\ifCLASSINFOpdf
\else
\fi
\begin{document}
%
\title{Deep Learning for Sampling from Arbitrary Probability Distributions}



\author{\IEEEauthorblockN{Felix Horger, Tobias W\"urfl, Vincent Christlein and Andreas Maier}
\IEEEauthorblockA{Pattern Recognition Lab of the Friedrich-Alexander-University Erlangen-Nuremberg\\
Email: felix.horger@fau.de, tobias.wuerfl@fau.de, vincent.christlein@fau.de and andreas.maier@fau.de\\
Homepage: https://www5.cs.fau.de}
}


\maketitle

\begin{abstract}
	This paper proposes a fully connected neural network model to map samples from a uniform distribution
	to samples of any explicitly known probability density function.
	During the training, the Jensen-Shannon divergence between the distribution of the model's output
	and the target distribution is minimized.\par
	We experimentally demonstrate that our model converges towards the desired state.
	It provides an alternative to existing sampling methods such as inversion sampling, rejection sampling,
	Gaussian mixture models and Markov-Chain-Monte-Carlo. Our model has high sampling efficiency
	and is easily applied to any probability distribution, without the need of further analytical or numerical calculations.
	It can produce correlated samples, such that the output distribution converges faster towards the target than for independent
	samples. But it is also able to produce independent samples, if single values are fed into the network and the input values are
	independent as well.\par
	We focus on one-dimensional sampling, but additionally illustrate a two-dimensional
	example with a target distribution of dependent variables.
\end{abstract}


%
\IEEEpeerreviewmaketitle

\section{Introduction}
Sampling from a given probability density function (PDF) is required in various applications,
e.g. computer graphics, Monte-Carlo- or physical simulations. Existing
methods for generating such samples include inversion sampling, rejection sampling,
Gaussian Mixture or Markov-Chain-Monte-Carlo methods. In any case, random samples of simple
distributions, such as a uniform distribution, are used to generate samples of the desired distribution.\par
Here we examine how a fully connected neural network (FCNN) model performs on this task. Our model is constructed to
map an input vector consisting of $n$ samples from a uniform distribution to an output vector with the same dimension.
The target distribution has to be explicitly known and normalized.\par
Such a model yields a high sampling efficiency since $n$ input samples are required to generate $n$ output samples. It is flexible towards the
choice of the target PDF and needs little manual effort.\par
This paper gives a brief overview of the mentioned sampling methods and compares them to our
FCNN model in one dimension, regarding the properties of the produced samples
and the effort, both computational and manual. We also elaborate an example of sampling from
two-dimensional PDFs.

\section{Conventional Sampling Methods}

\subsection{Inversion Sampling}
\label{inv_sampling}
This method provides a function, which maps samples from
an arbitrary distribution $\alpha(x)$ to samples from the target PDF $\rho(y)$.
Let $\varphi\!\!:x\mapsto y$ be this function, then, assuming that $\varphi$ is a
bijection, the differential equation
\begin{equation}
	\alpha(x)\,\textrm{d}x \>=\> \rho(\varphi(x))\,\textrm{d}\varphi
	\label{eq:diff_eq}
\end{equation}
holds. If $\alpha$ is the uniform distribution over $[0, 1]$, then $\varphi$ can be
determined to be equal to the inverse cumulative distribution function (CDF) corresponding to $\rho$.
In many cases, the CDF or its inverse might not have an analytical representation.
If the integration and inversion are performed numerically, the computational effort
increases and the quality of the samples decreases.
In contrast, if $\varphi$ is explicitly known, this method has high efficiency and produces samples with
exactly the desired properties. If the samples drawn from $\alpha$ are
independent, the same holds for the produced samples.\par
This method may be applied to higher dimensional PDFs, using either the separability
of PDFs of uncorrelated random variables or the Bayes' theorem for correlated random variables, to split
the sampling process into multiple one-dimensional sampling steps. For this approach all the
conditional one-dimensional PDFs have to be known \cite[526ff]{Bishop2006PRML}.

\begin{figure*}[!t]
\centering
\subfloat[Average of $500$ KDEs each from $500$ random values produced with the mixture of Gaussians method.]{
\resizebox{2.0in}{!}{\input{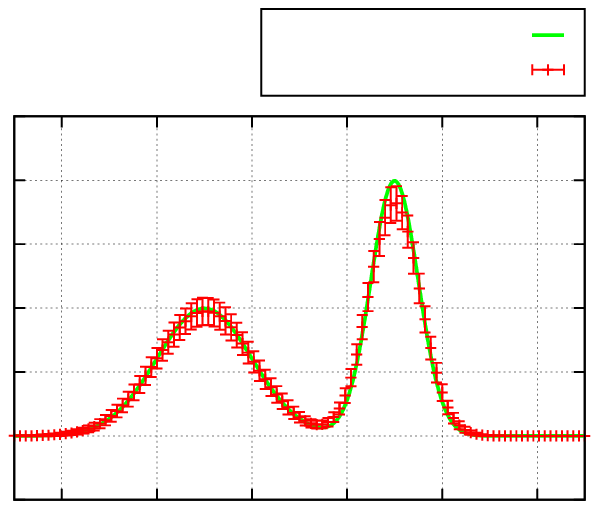}}
\label{fig:gmgmkde}}
\hfil
\subfloat[Mean of the KDE of $500$ output vectors, each consisting of $500$ elements. The training was performed
on $5\cdot10^{6}$ input vectors drawn from the uniform distribution over {$[-1, 1]$}.]{
\resizebox{2.0in}{!}{\input{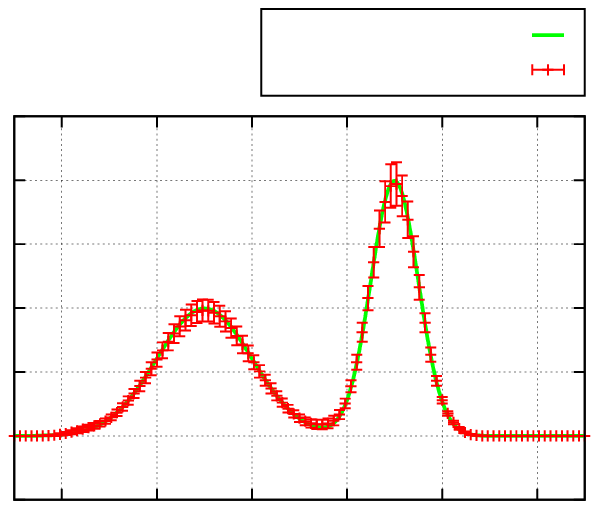}}
\label{fig:gm500to500kde}}
\hfil
\subfloat[Two randomly chosen elements of $500$ output vectors plotted against each other, showing the dependence of
the model's output values.]{
\resizebox{2.0in}{!}{\input{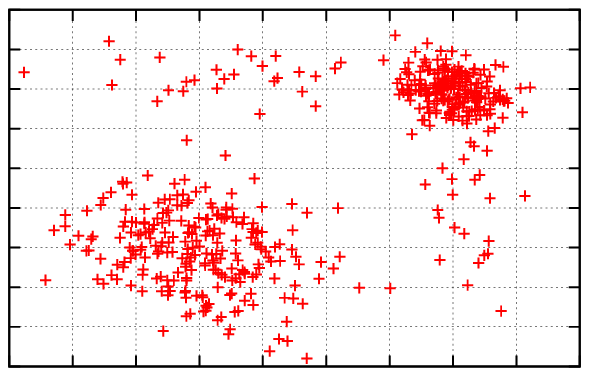}}
\label{fig:gmdependence}}
\caption{Comparison of the model to the mixture of Gaussians method on a bimodal Gaussian target, given by \eqref{eq:bimodaltarget}.}
\label{fig_double_bimodal}
\end{figure*}

\subsection{Rejection Sampling}
\label{rejectionsampling}
This method requires a proposal distribution $\beta(y)$ from which sampling can be performed.
Additionally, there has to exist a constant $c$ such that $c\cdot\beta(y)\geq \rho(y) \,\,\, \forall y$.\par
The procedure starts with drawing a sample $y$ from $\beta$ and another sample $r$ from the uniform
distribution $R$ over $[0, c\cdot\beta(y)]$. If $r < \rho(y)$, then $y$ is a valid sample from
$\rho$, otherwise $y$ is rejected. This process is continued until enough valid samples have been generated.\par
The major disadvantage of this method is that sampling efficiency depends on how close the proposal
distribution lies to the target distribution.
Besides that, the produced samples are independent samples from $\rho$, if the samples
drawn from $\beta$ and $R$ are independent. Another advantage is that the target does
not have to be normalized \cite[528ff]{Bishop2006PRML}.

\subsection{Mixture of Gaussians}
The quality of the samples produces by this method depends on how well a suitable sum of Gaussians
approximates the target. Samples from the approximation can be obtained by randomly choosing a Gaussian
mode from the sum with probability proportional to its
weight and generating a sample from it.\par
This method is easy to perform since sampling from a Gaussian PDF can be done using inversion sampling
and if the Gaussian samples are independent, the generated samples are independent as well \cite[110ff]{Bishop2006PRML}.

\subsection{Markov-Chain-Monte-Carlo}
The aim of this method is to construct a Markov chain with a stationary distribution
equal to the target. The Metropolis-Hastings algorithm is a commonly used method of
doing so. An initial sample $y_0$ is used to propose a possible next sample $y'$, drawn
from an arbitrary conditional distribution $q(y' | y_0)$. The sample is
accepted as $y_1$, if
\begin{equation}
	r \leq \min\left\{ 1,\, \frac{\rho(y')}{\rho(y_0)}\cdot\frac{q(y_0 | y')}{q(y' | y_0)} \right\}
	\label{eq:mcmc_rejection}
\end{equation}
where $r$ is drawn from the uniform distribution over $[0, 1]$. If $y'$ is rejected,
$y_1$ is equal to $y_{1-1} = y_0$. This process is continued, until enough samples were obtained.\par
The proposal distribution $q(y' | y)$ has a great impact on the convergence of the distribution of
the samples towards the target distribution. Consider for example a target with two modes placed distant to each other and a narrow proposal
distribution located at $y$. It is very unlikely to switch between the modes, leading
to slow convergence. Further, the initial sample $y_0$ has impact on the convergence: if $\rho(y_0)$ is
small, it might need some time to reach areas of higher probability (``burn in''). Taking into account
that the next sample is generated using the last one yields that the samples depend on each other \cite[539ff]{Bishop2006PRML}.\par
There is a possibility to link Markov-chain-Monte-Carlo methods and generative adversarial networks \cite{Goodfellow2014GAN}
in order to produce random numbers \cite{SongNNMCMC17}. The generator is used as the transition kernel
of the Markov chain if samples from the target distribution are accessible during the training. If
this is not the case, the generator is trained to propose samples $y'$. Again, the proposal
distribution has a great impact on the convergence and also the correlation of samples, thus this method
is optimizing the step of proposing, leading to fast convergence and low correlation.

\begin{figure*}[t]
\centering
\subfloat[Histogram of $10^4$ output values from the model.]{
\resizebox{2.5in}{!}{\input{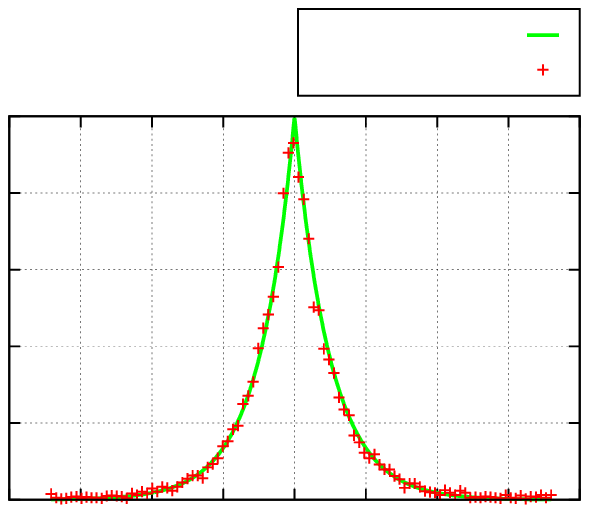}}
\label{fig:1to1expabs}}
\hfil
\subfloat[Histogram of $10^4$ random values produced with the inversion sampling method.]{
\resizebox{2.5in}{!}{\input{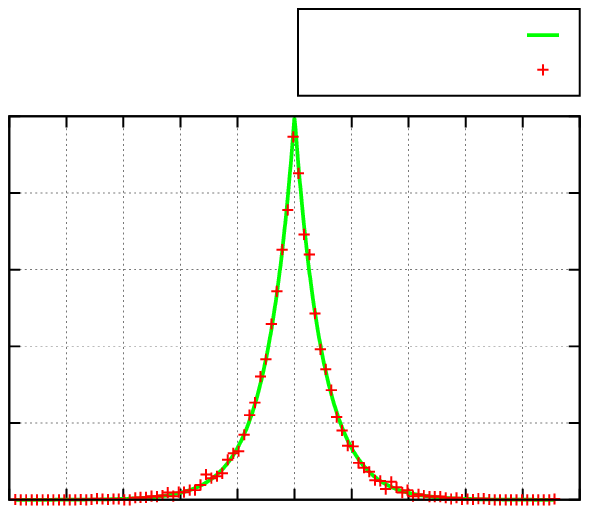}}
\label{fig:invsampleexpabs}}
\caption{Comparison of the model to the inversion sampling method on an exponential target given by \eqref{eq:expabs_target}.}
\label{fig_double_expabs}
\end{figure*}

\section{Sampling with a Fully Connected Neural Network}

\subsection{Concept and Setup}

Our model is a simple FCNN, which is able to map an input vector
consisting of $n$ samples from the input distribution to an output vector of the same dimension. The target
distribution has to be explicitly known. The term ``sample'' refers in this context only to an element of any
in- or output vector and is not to be understood as a ``sample from the training set''.\par
The model has $n$ units in any layer and exponential linear unit (ELU) activation \cite{ClevertUH2015ELU} in each but the last layer.
A number of layers equal to ten has proven to lead to good results. The input dimension is $n = 500$, limited by the resources
of the used hardware.\par 
The ADAM optimizer \cite{KingmaB2014ADAM} was used for the training process, the weights were initialized
using the Xavier Glorot uniform distribution \cite{Glorot2010DTDNN} and the biases were set to zero.
The models were implemented using Python and Keras \cite{Chollet2015Keras} with TensorFlow \cite{tensorflow2015-whitepaper} backend.\par
The loss-function of the model consists of three parts. The kernel density estimation (KDE) \cite{Silverman1986KDE} of each
output vector in a mini-batch is compared to the target PDF. Additionally, the $i$-th element from each output vector in a mini-batch is extracted, treated as a set samples and
its KDE is compared to the target. This performed for all $i$. It promotes diversity, otherwise the model produces the same output
vector with the correct distribution for any input.\par
The comparison of the KDE and the target distribution may be done using the mean-squared-error or the Jensen-Shannon-divergence
\cite{Endres2003JSD}, it was empirically found that the latter yields better results for most cases. The Jensen-Shannon-divergence
\begin{equation}
	D_{JS}(p,\, q) = \frac{1}{2}\,\int\limits_{\mathbb{R}} \left[
	p(y)\,\log\frac{p(y)}{q(y)} +
	q(y)\,\log\frac{q(y)}{p(y)}
	\right] \,\text{d}y
	\label{eq:jsd}
\end{equation}
between the PDFs $p$ and $q$ measures their similarity.
The above integral has to approximated numerically, which is possible due to the properties of $p$ and $q$.
It is calculated on a finite set of discrete values. The third part of the loss-function to confines the output values in between the
borders of this set using a ``potential well'', which was chosen to have linearly increasing sides.\par
The uniform distribution over $[-1,1]$ was used to generate input samples. Any input distribution with zero
mean leads to equal results, other distributions perform significantly worse. This is caused by the
internal covariate shift, since no batch normalization was used for our model \cite{Ioffe2015BN}.\par

\begin{figure*}[!t]
\centering
\subfloat[Histogram of $10^4$ output values from our model.]{
\resizebox{2.0in}{!}{\input{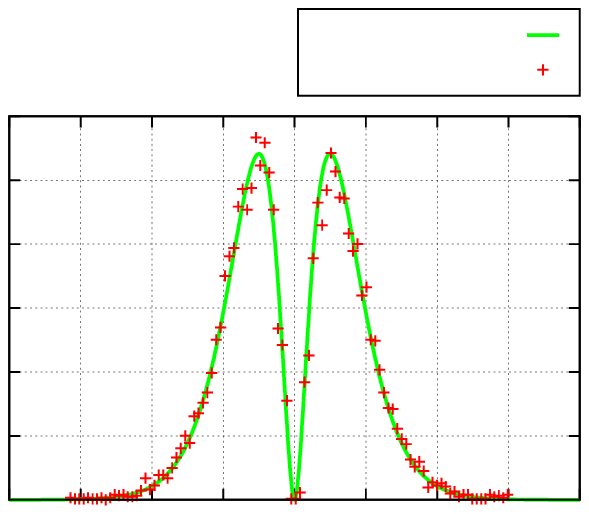}}
\label{fig:1to1y2expabs}}
\hfil
\subfloat[Histogram of $10^4$ random values produced with the rejection sampling method.]{
\resizebox{2.0in}{!}{\input{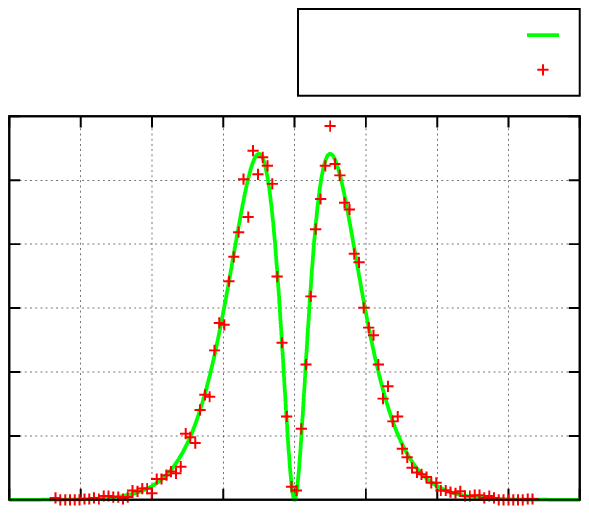}}
\label{fig:rejsamply2expabs}}
\hfil
\subfloat[Histogram of $10^4$ random values produced with the Metropolis-Hastings algorithm.]{
\resizebox{2.0in}{!}{\input{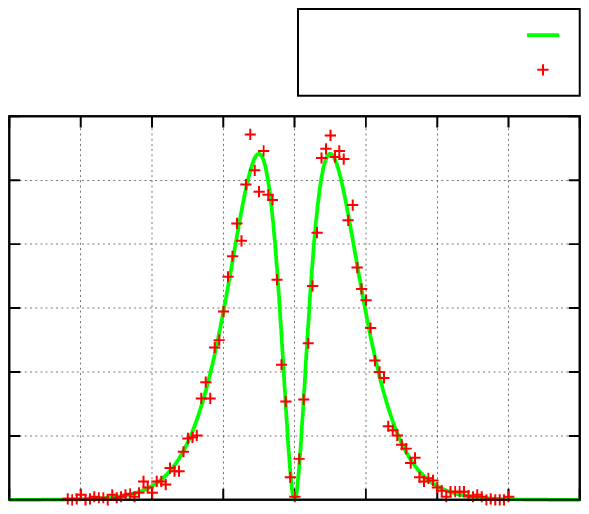}}
\label{fig:mcmcy2expabs}}
\caption{Comparison of our model to rejection sampling and the Metropolis-Hastings algorithm on the target given by \eqref{eq:y2expabsy}.}
\label{fig_double_y2expabs}
\end{figure*}
%

\begin{figure*}[!t]
\centering
\subfloat[2D KDE of $10^4$ output values from the model.]{
\resizebox{2.5in}{!}{\input{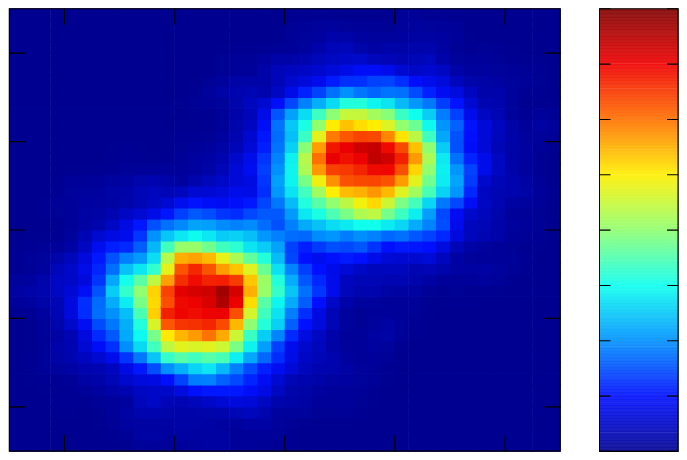}}
\label{fig:2d_mean}}
\hfil
\subfloat[The two-dimensional bimodal Gaussian target PDF.]{
\resizebox{2.5in}{!}{\input{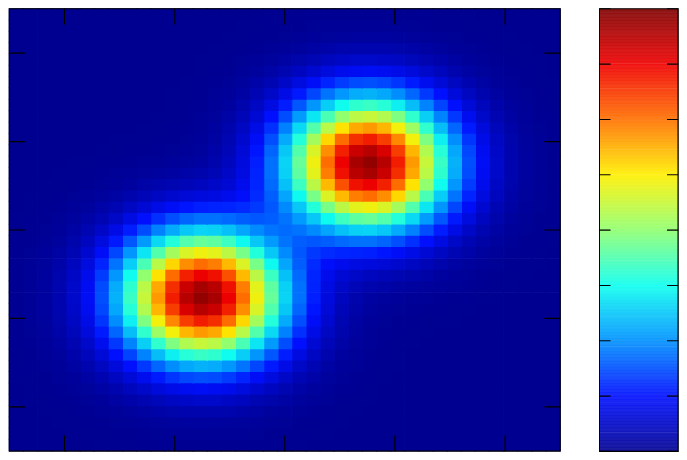}}
\label{fig:2d_target}}
\caption{Comparison of the final KDE of our model's output values to the two-dimensional bimodal Gaussian target.}
\label{fig_double_2d}
\end{figure*}

\subsection{Results}
Using this setup, our model is able to produce samples of the given target distribution, i.e. the kernel density
estimation of the output values converges towards the target PDF. The input dimension is equal to $500$ and the model has ten layers.
The target is
\begin{equation}
	\rho(y) \;=\; \frac{2\cdot\exp\left[2\cdot(x-1)^2\right] + \exp\left[\frac{1}{2}\,(x+3)^2\right]}{2\sqrt{2\pi}}
\label{eq:bimodaltarget}
\end{equation}
which is a bimodal asymmetric Gaussian.
The resulting KDE of the output values is shown in Figure \ref{fig:gm500to500kde}.\par
Since $500$ values are fed at once into the network, it may happen that the output values depend on each other.
This property allows the KDE of the output values to lie closer to the target PDF as if the values were drawn
independently, e.g. using the mixture of Gaussians method (see Figure \ref{fig:gmgmkde}). Consider that the
KDE is calculated of only $500$ values, which are too few for a reasonable estimation of the underlying PDF. In fact,
the model makes the output values interdependent, in order to overcome this issue. The dependence can
be more clearly seen in Figure \ref{fig:gmdependence}, where two randomly chosen elements of the output vectors
are plotted against each other. If they are independent, there would be peaks at $(-3,\, 1)$ and $(1,\, -3)$, too.\par
If independent samples are required, the input dimension may be reduced to one.
Such setup introduces no further correlation and thus the output values are independent if the input values are independent as well.
The model used in this scenario has $500$ units per layer and ten layers. We can shown experimentally that the model
with input dimension one represents the mapping function given by the inversion sampling method. This is not surprising because
the differential equation \eqref{eq:diff_eq} has a single solution on every finite subset of $\mathbb{R}$,
given any boundary condition (Picard-Lindel\"of theorem, note that there exist other mapping functions, which are
not bijections, see Chapter \ref{inv_sampling}).\par
The model with input dimension one was compared to the inversion sampling method for the target
\begin{equation}
	\rho(y) = \frac{1}{2}\exp(-|y|)
	\label{eq:expabs_target}
\end{equation}
an exponential distribution with extended domain $D = \mathbb{R}$.
The training was performed on $10^7$ input values and the histogram displayed in Figure \ref{fig:1to1expabs} was calculated
out of $10^4$ output values. In contrast, Figure \ref{fig:invsampleexpabs} shows the histogram of just as many random values
obtained from the inversion sampling method.\par
Comparing these two yields no difference except that in Figure \ref{fig:invsampleexpabs} higher values occur. This may have been
caused by the numerical precision and a more complicated fitting at the borders, since the model is represented by a continuous
function, but the inverse CDF diverges at zero and one.
This model was further compared to the rejection sampling method with a target
\begin{equation}
	\rho(y) \propto y^2\cdot\exp(-b|y|) \hspace{0.5cm} \text{with} \hspace{0.5cm} b > 0
	\label{eq:y2expabsy}
\end{equation}
that has an inverse CDF with no analytical representation.
Figure \ref{fig:1to1y2expabs} shows the resulting histogram of $10^4$ output values after a training on $10^7$ input values.
Comparatively, the histogram of the same amount of values produced with the rejection sampling method is displayed in Figure
\ref{fig:rejsamply2expabs}. The proposal distribution was chosen to be equal to \eqref{eq:expabs_target}. This is not the
best choice since it has its maximum where the target is zero. Note that a bimodal Gaussian proposal can not be used since
no constant $c$ fulfills the condition given in Chapter \ref{rejectionsampling}. But a poor choice of the proposal distribution
does not affect the quality of the samples, only the computational effort.\par
Comparing Figure \ref{fig:1to1y2expabs} and \ref{fig:rejsamply2expabs} yields that the histogram of the samples produced by our model 
approximates the target as well as samples produced with the rejection sampling method.\par
The same target was used for the Metropolis-Hastings algorithm, the proposal distribution was chosen to be a Gaussian with standard deviation
$0.5$ located at the current sample. As in Figure \ref{fig:mcmcy2expabs} depicted, the histogram of the samples approximates the target and
the goodness of the fit is comparable to Figures \ref{fig:1to1y2expabs} and \ref{fig:rejsamply2expabs}.\par
Further, the model is able to sample from two-dimensional PDFs of dependent variables. The model was trained on $2\cdot10^8$ input samples
for a 2D bimodal Gaussian target with peaks at $(\pm1.5, \pm1.5)$ and variances of one in each direction.
The result together with the target is depicted in Figure \ref{fig_double_2d}.\par
The computational effort for the comparison of the estimation of output distribution and the target scales exponentially with the dimension.
So there is a trade-off between training time and correct PDF-estimation. A possible solution is to manually split the PDF into its conditional
one-dimensional parts and train a separate model for each dimension.

\section{Conclusion}
Summarizing this paper, our FCNN model is able to sample from any target PDF.\par
The presented findings show that our model produces results with a goodness of the fit comparable to
any existing sampling method. The quality of the approximation can be tuned with the size of the model
and the duration of the training.\par
In order to apply our model, a normalized target PDF is required, in contrast to the Metropolis-Hastings
algorithm or the rejection sampling method. On the other side, no proposal distribution, constant $c$ (rejection sampling) or
location of Gaussian modes (Gaussian mixture) has to be determined for our model. The only parameter of our model that has to be
adjusted by hand is the width of the kernel function for the KDE of the output values.\par
Our model has the highest possible sampling efficiency equal to the inversion sampling method. Whereas the other described methods
transform multiple samples into a single one. Especially the rejection sampling method may have low sampling efficiency.\par
Another important advantage is the flexibility towards the choice of the target.
Compared to inversion sampling, no integration or inversion is required, neither analytical nor numerical. Our model with input dimension one
tries to represent the inverse CDF. This raises the question if the inversion may not simply be performed numerically from the beginning. Instead of using our model,
the fit can be performed in any other way. But an important advantage of FCNN models is that they are universal function
approximators. It is neither required to choose how to proceed with the inversion, nor being bound to an approximation with Gaussian modes.\par
For high input dimensions, our model is able to generate dependent samples such that their distribution converges faster towards the target than the distribution of
independent samples would do. If the input dimension is set to one, our model is able to produce independent samples if the input values are
independent as well. This makes it more attractive than the Metropolis-Hastings algorithm, which produces highly dependent values and may have slow convergence.\par
Sampling from two-dimensional PDFs of dependent variables is also possible, but the curse of dimensionality has not yet been overcome. 
Splitting the target into conditional one-dimensional distributions using Bayes' theorem is a possible solution.
\newpage

\if (0)
Each of the conventional sampling methods has its advantages and disadvantages.  our model
produces results with a similar goodness of the fit. The quality of the approximation can be specified by the size of the model and the duration of the training.
It can easily be deployed for any normalized target PDF since only the target PDF itself has to be explicitly known.\par
Our model is also able to produce independent samples, when using an input dimension equal to one. This circumstance makes it more attractive than the
Metropolis-Hastings algorithm, which produces highly dependent values. If the input dimension is chosen higher, our model produces dependent output values.
Their distribution converges faster towards the target than the distribution of independent random values.\par
Compared to inversion sampling, our model achieves the same sampling efficiency, but no integration or inversion is needed, neither analytical nor numerical.
The point that the model tries to represent
the inverse CDF, raises the question if the inversion may not simply be performed numerically from the beginning. Instead of using our model,
the fit can be performed in any other way, e.g. see \cite{OlverFITS}. But an important advantage of FCNN models is that they are universal function
approximators and they are easily applied to any target, without the need of choosing how to proceed with the inversion.\par
Regarding rejection sampling and the Metropolis-Hastings algorithm, no proposal distribution or constant $c$ has to be determined, which may
influence the computational effort or quality of the samples. Our model sampling efficiency is higher, but a disadvantage of the presented model
is that the target has to be normalized, in contrast to the rejection sampling method.\par
When approximating any PDF, the mixture of Gaussians method may be easier to apply, but our model is more flexible, since it is not bound to
use Gaussian peaks for the approximation.\par

Sampling from two-dimensional PDFs of dependent variables is also possible, but the curse of dimensionality has not yet been overcome. 
Splitting the target into conditional one-dimensional distributions using Bayes' theorem is a possible solution.
\newpage
\fi





\bibliographystyle{IEEEtran}
\bibliography{bibliography}
%



\end{document}